# scientific reports

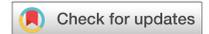

OPEN

# An efficient plant disease detection using transfer learning approach

Bosubabu Sambana[1], Hillary Sunday Nnadi[2], Mohd Anas Wajid[3], Nwosu Ogochukwu Fidelia[2✉], Claudia Camacho-Zuñiga[3], Henry Dozie Ajuzie[4] & Edeh Michael Onyema[5,6]

Plant diseases pose significant challenges to farmers and the agricultural sector at large. However, early detection of plant diseases is crucial to mitigating their effects and preventing widespread damage, as outbreaks can severely impact the productivity and quality of crops. With advancements in technology, there are increasing opportunities for automating the monitoring and detection of disease outbreaks in plants. This study proposed a system designed to identify and monitor plant diseases using a transfer learning approach. Specifically, the study utilizes YOLOv7 and YOLOv8, two state-of-the-art models in the field of object detection. By fine-tuning these models on a dataset of plant leaf images, the system is able to accurately detect the presence of Bacteria, Fungi and Viral diseases such as Powdery Mildew, Angular Leaf Spot, Early blight and Tomato mosaic virus. The model's performance was evaluated using several metrics, including mean Average Precision (mAP), F1-score, Precision, and Recall, yielding values of 91.05, 89.40, 91.22, and 87.66, respectively. The result demonstrates the superior effectiveness and efficiency of YOLOv8 compared to other object detection methods, highlighting its potential for use in modern agricultural practices. The approach provides a scalable, automated solution for early any plant disease detection, contributing to enhanced crop yield, reduced reliance on manual monitoring, and supporting sustainable agricultural practices.

The global human population is undergoing rapid growth, leading to substantial impacts and posing significant challenges to humanity. Among these challenges, the scarcity of food has emerged as a critical issue, primarily stemming from the inadequate availability of essential resources like arable land, water, and labor. Several factors contribute to this problem, including climatic conditions that lead to crop failures, infestations by plant pests, and the intrusion of pathogens, all of which collectively result in decreased crop production. According to the international center for agricultural research (ICAR), an estimated annual loss of over 35% in agricultural productivity can be attributed to pest and disease-related factors[1]. The increasing incidence of insect infestations and crop diseases is creating a precarious situation for global food security. Furthermore, the ramifications of these plant diseases extend beyond food security, encompassing far-reaching impacts on the economy, society, and the environment.

Plant diseases have a negative impact on agricultural productivity, leading to crop losses and reduced nutritional value. Timely diagnosis and treatment are crucial to prevent the further spread of the disease and reduce its impact on crop yields. Early warning and forecasting play essential roles in successful plant disease management, facilitating effective monitoring and intervention in agricultural production. Currently, visual assessments conducted by experienced farmers are the most common method of detecting plant diseases in rural regions, which often require consultation with specialists. However, this approach may be economically not feasible for large farms, and farmers in remote areas may face challenges accessing specialized expertise, resulting in high costs and time consumption. Hence, there is a need for quick, automated, cost-effective, and accurate methods for identifying plant diseases[2]. Researchers are exploring the application of computer vision techniques for scalable and economical plant disease diagnosis. Deep convolutional neural networks (CNNs) have made significant advancements in the field of computer vision[3]and present a promising approach for achieving both rapid and accurate diagnosis of plant diseases. Once trained, these models can quickly classify images, making them suitable for mobile applications[4]. Combining both computer vision techniques associated with transfer learning methods results in creating new opportunities for resolving various agricultural problems related to plant disease classification and detection ensuring better crop yields as outcomes which is done by effectively managing plant diseases.

[1]Department of Computer Science and Engineering (Data Science), School of Computing, Mohan Babu University, Tirupathi, Andhra Pradesh, India. [2]Department of Computer and Robotics Education, University of Nigeria, Nsukka, Nigeria. [3]Institute for the Future of Education Tecnologico de Monterrey, Monterrey, Mexico. [4]Department of Educational Foundations, University of Nigeria, Nsukka, Nigeria. [5]Department of Mathematics and Computer Science, Coal City University, Enugu, Nigeria. [6]Adjunct Faculty, Saveetha School of Engineering, Saveetha Institute of Medical and Technical Sciences, Chennai 602105, India. ✉email: fidelia.nwosu@unn.edu.ng

The growing global demand for food, coupled with the challenges posed by climate change, has made plant disease management a critical component of agricultural productivity. Plant diseases can devastate crops, leading to significant losses in yield and quality, which in turn threatens food security and the livelihood of farmers. Early detection of plant diseases is essential for effective intervention and management. Traditional methods of disease identification often rely on manual inspection by experts, which can be time-consuming and prone to human error. Additionally, these methods may not be scalable for large-scale agricultural operations. As such, there is a need for automated systems that can detect plant diseases rapidly and accurately, offering timely solutions for farmers to address potential outbreaks before they cause widespread damage.

In recent years, advancements in machine learning and computer vision have provided innovative solutions for plant disease detection. Among these, deep learning techniques, particularly convolutional neural networks (CNNs), have shown great promise due to their ability to learn complex patterns from large datasets. Transfer learning, a technique that adapts pre-trained models for new tasks, has further enhanced the efficiency and accuracy of these systems by reducing the need for large labeled datasets. This study explores the application of transfer learning with modern object detection models, specifically YOLOv7 and YOLOv8, for plant disease detection. These models, known for their speed and accuracy in detecting objects in images, are fine-tuned to classify and identify diseases in plant leaves. By leveraging these advanced techniques, the proposed system aims to provide an efficient, scalable, and reliable solution for early plant disease detection, ultimately improving crop management and supporting sustainable agricultural practices. The study introduces the utilization of object detection approaches, specifically YOLOv7 and YOLOv8, to tackle the issue of identifying plant diseases under intricate circumstances. The evaluation employed the Detecting Diseases Dataset, curated in an environment lacking strict controls. To determine the superior model, the assessment considered metrics like mean average precision (mAP), precision, recall, and F1-score.This study involves the training of object detection models using the TensorFlow and Keras libraries. Google Colab, a notebook service similar to Jupyter, was employed for its provision of parallel computing resources dedicated to training machine learning models. Notably, Google Colab grants complimentary access to Graphics Processing Units (GPUs) to expedite computations. The GPU utilized in this context is the Tesla T4, equipped with 12.68GB of memory and 78.19GB of disk space.

The study focuses on four key plant diseases that have a considerable impact on tomato crop health and yield: powdery mildew, angular leaf spot, early blight, and tomato mosaic virus. Powdery Mildew is caused by a variety of fungal infections, resulting in white, powdery growth on leaf surfaces, lowering photosynthetic efficiency and damaging plants. Furthermore, Angular Leaf Spot predominantly affects the plant and causes distinctive angular lesions on leaves and fruit, significantly reducing plant vigour and marketability. Similarly, Early Blight is a widespread fungal disease that produces concentric ring patterns on leaves and stems, resulting in premature defoliation and lower yields. Tomato Mosaic Virus is a highly transmissible viral infection that causes mottling, chromatic aberration and slowed growth in afflicted plants, causing a significant danger to production. Early detection and management of these diseases is critical for creating sustainable prevention techniques and maintaining environmentally friendly agricultural methods.

## Literature review

Recent advancements in precision agricultural technologies[3] have led to a substantial increase in crop production. However, this gain in crop yield has raised concerns about declining product quality[5]. Agricultural product quality is negatively impacted by plant diseases[6]. Traditional methods[7] involve meticulous plant species examination, yet these procedures are resource-intensive in terms of time and money. The progress of IoT[8] and machine learning has emphasized the necessity of digitizing plant disease detection[7].

As far as our knowledge goes, the PlantDoc dataset[4] and the PlantVillage dataset (PVD)[9] are the sole publicly accessible databases for plant disease identification. The PlantVillage dataset utilized GoogleNet and AlexNet, achieving a remarkable 99.35% accuracy. Nevertheless, its images were captured in controlled lab environments, possibly limiting its applicability in real-world agricultural contexts. On the other hand, the PlantDoc dataset contains real-time images of both diseased and healthy plants.In[10], introduced a YOLOv5-based deep learning method for early detection of bacterial spot disease in bell pepper plants, achieving a robust mean average precision (mAP) score of 90.7%. This model holds the potential for assisting farmers in timely disease identification. Similarly some authors used deep learning to handle the specific problem[11,12], and some worked on the same problem[13], addressed rice leaf diseases using YOLOv5, outperforming other methods in object detection accuracy. Their model achieved recall, precision, and mAP scores of 0.94, 0.83, and 0.62, respectively, for four distinct rice leaf diseases, contributing to improved crop quality and yield.Because it offers a framework to address the inherent ambiguities and difficulties connected with the process, neurophysiology plays a critical role in the identification of plant diseases. Neutron spectroscopy facilitates the representation and analysis of unknown or unclear elements, such as pathogen interactions, genetic variability, and environmental conditions, that impact disease manifestation in the context of plant disease detection[14,15]. In[15], a unique plant disease diagnosis strategy is formulated, training deep learning models on disease patch photos irrespective of the crop being diagnosed, showcasing improved performance compared to conventional crop-disease pair-based strategies. In another study[16], the Yolov5 deep learning model is utilized to detect rice leaf diseases. The model is trained on a dataset encompassing images of four distinct rice leaf diseases, achieving mAP, recall, and precision scores of 0.62, 0.94, and 0.83, respectively.In order to create complete models for illness diagnosis and prediction, neurosophic techniques allow the integration of uncertain data sources, such as sensor measurements, satellite images, and expert knowledge[17]. Neutronophy, which embraces ambiguity and uncertainty, makes it easier to create reliable decision support systems that can detect and treat plant diseases, improving crop health and agricultural output[18].

The YOLOv7[19] and YOLOv8[8] models emerge as the latest object detection detectors. These networks employ trainable bag-of-freebies to enhance accuracy without increasing inference costs. Moreover, the target detector employs extend and compound scaling to significantly enhance detection time by efficiently reducing parameters and computations[19]. As of now, YOLOv7 and YOLOv8 stand as the cutting-edge detectors yet to be employed for plant disease detection. Thus, the present study utilizes YOLOv7 and YOLOv8 to detect plant diseases, yielding unprecedented accuracy results in plant disease detection.

The proposed strategy addresses numerous obstacles in diseases of plants detection, such as constrained labeled data, significant variation across plant genera, and the need for real-time, Resource-effective solutions, by exploiting pre-trained models that have previously acquired extensive representations of features from large, diverse datasets. By fine-tuning these models on plant disease datasets, transfer learning allows for the deployment of lightweight, efficient architectures that perform well across many datasets. This strategy shortens training time, decreases processing needs, and assures that the model extends well to new types of plants and diseases, even with little and diverse datasets.

It is worth noting that previous research lack scalability due to overfitting on limited datasets or heavy model designs unsuited for practical deployment, as well as variety and model generalization. The proposed transfer learning strategy tries to address these challenges by constructing a lightweight, efficient model architecture that performs well across various datasets. We believe that these improvements better justify the reason for our effort and articulate its benefit for the area.

### Traditional methods for plant disease detection

Traditional plant disease detection methods primarily relied on visual inspection and expert knowledge to identify symptoms of diseases on plant leaves and crops. These techniques involved farmers or agricultural experts manually identifying visible signs of disease, such as discoloration, spots, and lesions, which could indicate a particular condition. While effective in certain contexts, these methods have several limitations, including subjectivity, time consumption, and the requirement for highly trained personnel. As a result, these methods are not easily scalable or reliable for large agricultural operations. Over time, researchers have attempted to automate the disease detection process using image processing techniques, such as color and texture analysis, edge detection, and pattern recognition. However, these methods were often limited by their inability to generalize across various environments, plant species, or diseases, highlighting the need for more advanced technologies.

### Recent deep learning approaches

In recent years, deep learning has revolutionized the field of plant disease detection by providing more accurate, efficient, and scalable solutions. Convolutional Neural Networks (CNNs), in particular, have gained popularity due to their powerful ability to learn complex features directly from raw image data, eliminating the need for manual feature extraction. Early studies in this area employed CNN models like AlexNet, VGG, and ResNet to classify plant diseases based on leaf images, showing significant improvements in detection accuracy over traditional methods. These models have been trained on large public datasets such as PlantVillage, which contains labeled images of various plant diseases. One of the key advancements in this field is the use of transfer learning, which allows deep learning models to leverage pre-trained weights from large datasets (such as ImageNet) to perform well even on smaller, specialized agricultural datasets. Transfer learning not only reduces the need for extensive labeled data but also enhances the model's generalization ability across different plant species and diseases. Additionally, modern object detection models, such as YOLO (You Only Look Once), have been employed for real-time plant disease detection, further improving both speed and accuracy by detecting diseases in images quickly and efficiently.

### Attention mechanisms in plant disease detection

Attention mechanisms, which were initially introduced in natural language processing tasks, have recently been applied to plant disease detection to enhance the performance of deep learning models. These mechanisms allow a model to focus on specific regions of interest in an image, particularly those that exhibit symptoms of disease. In plant disease detection, attention mechanisms can help the model identify subtle patterns or anomalies in the plant leaves that may be indicative of disease. By highlighting the most relevant areas, attention modules improve the model's ability to differentiate between healthy and diseased plants, leading to better classification accuracy. Furthermore, attention mechanisms have been integrated into CNN architectures, enhancing the feature extraction process. More advanced attention models, such as Transformer-based architectures, have shown promise in improving the model's ability to handle complex disease patterns and variations in plant images. These attention-driven approaches have demonstrated a higher level of precision in detecting plant diseases, as they enable models to concentrate computational resources on the most informative parts of an image, thus improving overall detection performance.

## Methodology

The methodology employed in this study for plant leaf disease detection using transfer learning involves several key steps. First, the input data, consisting of leaf images with annotations, is read and preprocessed to ensure its suitability for training. Once the dataset was prepared, the data was split into a training set comprising 80% of the samples and a testing set containing the remaining 20%. Then, an object detection model is selected as the base model for training. The Hyperparameter tuning is performed on the model using the Detecting Diseases dataset to adapt it specifically for disease detection. Finally, the performance of the trained model is evaluated to assess its accuracy and effectiveness in detecting diseases in plant leaves. The process of disease detection is visualized in (Fig. 1).

## Detecting diseases dataset

The Detecting Diseases dataset[20] was created by roboflow.com on Sep 2, 2022. It consists of 5494 images from 3 plant species and it is divided into 12 diseased classes. There are several bacterial, fungal, and viral illnesses in the diseased classes that affect food crops including Beans, Strawberry and Tomato. The diseases of these crops are Angular Leaf spot, Anthracnose Fruit Rot, Blossom Blight, Gray Mold, Leaf Spot, Powdery Mildew Fruit, Powdery Mildew Leaf, Leaf Mold, Spider Mites, ALS, Bean Rust.

## Collection of plant material

Collection of plant material complies with relevant institutional, national, and international guidelines and legislation We adopted the format proposed by Hildreth et al.[21].

## You only look only once (YOLOv7)

YOLOv7[19] is a recently introduced model that follows the previous version, YOLOv6[22]. It offers significant advancements in object detection performance without incurring additional inference and computational costs. Compared tooother popular object detectors, YOLOv7 surpasses them by reducing approximately50% of the computation needed and 40% of the parameters for state-of-the-art object detection algorithms. This reduction allows for faster inferences by maintaining the accuracy of detection.

YOLOv7 presents an improved and efficient network architecture that incorporates an effective feature extraction method, resulting in enhancing the performance of object recognition. Additionally, it makes use of a steady loss function and improves the labeling process and increasing the effectiveness of model training. The enhancements contribute to the overall effectiveness of YOLOv7 in object detection tasks. As a result, YOLOv7 achieves better detection results (Fig. 2) with significantly less computational hardware requirements compared to other deep learning models. Additionally, it can be trained more quickly on small datasets without relying on pre-trained weights. YOLO models often have the capacity to identify and classify objects concurrently by just looking at the input image or video once. This approach is the reason behind the algorithm's name, "You Look Only Once". The YOLOv7 model incorporates various strategies to strike a good balance between detection effectiveness and accuracy. These techniques include model scaling for concatenation-based models[20], E-ELAN (Extended efficient layer aggregation networks)[17], and model re-parameterization[23]. By integrating these techniques, YOLOv7achieves improved performance.

## You only look only once (YOLOv8)

YOLOv8[8] is the most recent detection model in the YOLO family, which is noted for its object-detecting capabilities. The architecture is similar to YOLOv7 in that it has a backbone, head, and neck with improved convolution layers and detection head which makes this as ideal choice for real-time plant disease detection. YOLOv8 uses CSPDarknet53[24] as its backbone, a deep neural network that extracts features at multiple resolutions (scales) by progressively down-sampling the input image (Fig. 3). The feature maps produced at different resolutions contain information about objects at different scales in the image and different levels of detailing and abstraction. YOLOv8 can incorporate different feature maps at different scales to learn about object shapes and textures, which helps it achieve high accuracy inmost object detection tasks. YOLOv8 backbone consists of four sections, each with a single convolution followed by a c2f module. The c2f module is a new introduction to CSPDarknet53. The module comprises splits where one end goes through a bottleneck module (Two 3×3 convolutions) with residual connections. The bottleneck module output is further split N times where N corresponds to the YOLOv8model size. These splits are finally concatenated and passed through one final convolution layer associated with the activation function.

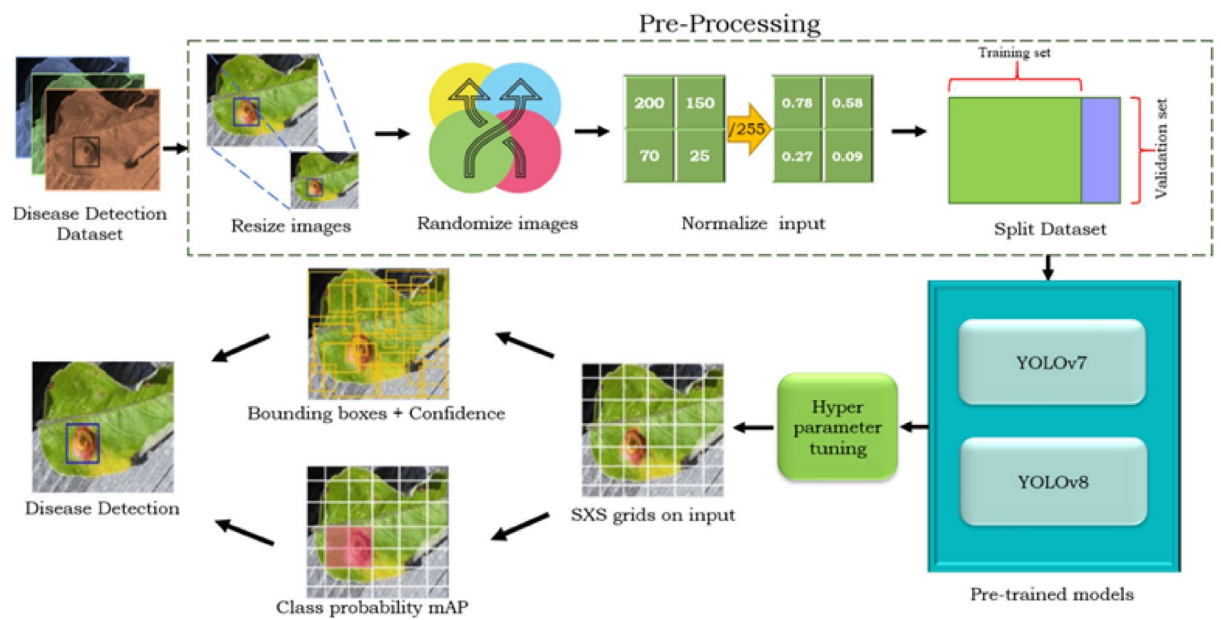

**Fig. 1.** Flowchart for plant disease detection.

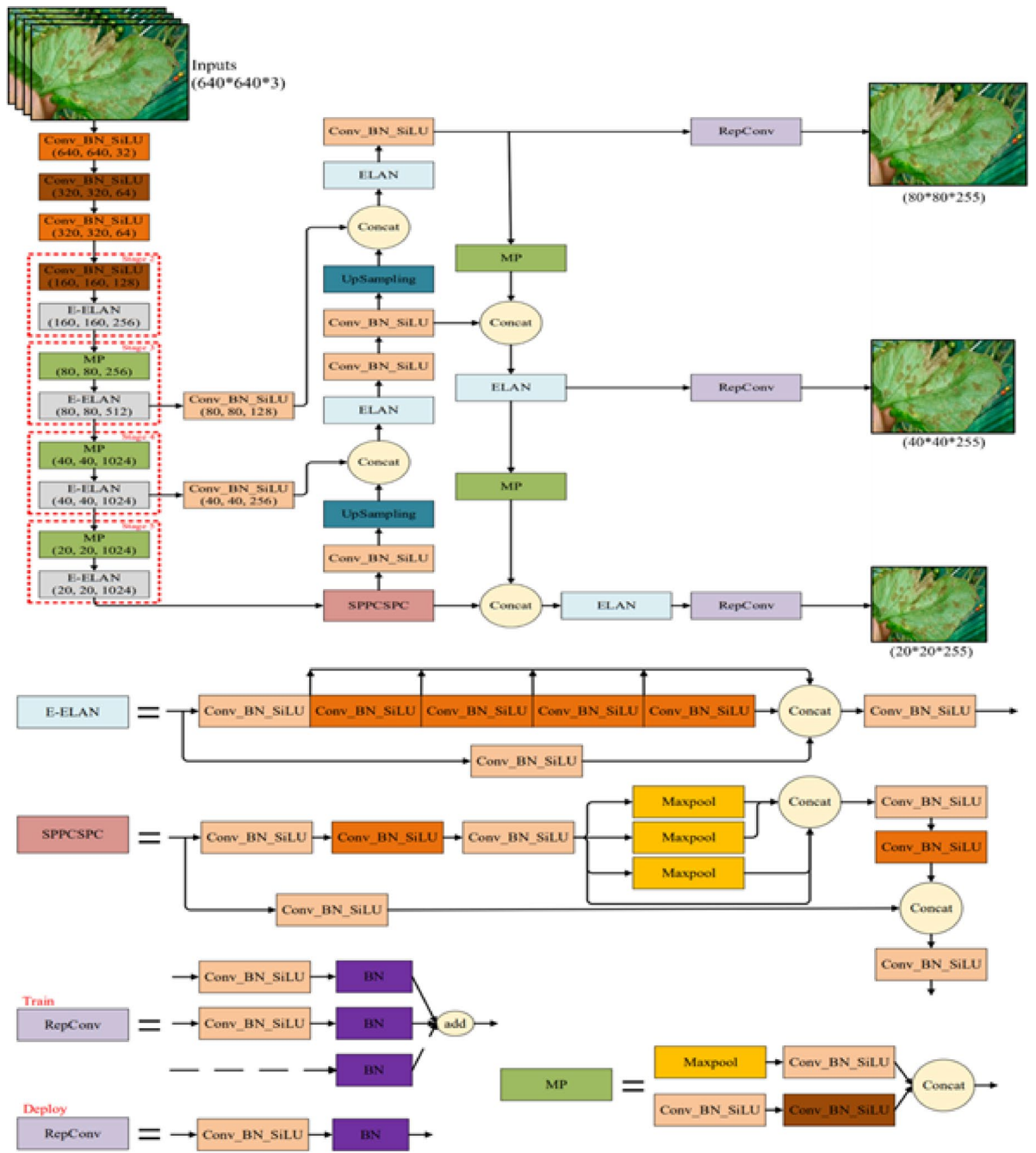

**Fig. 2.** Architecture of YOLOv7.

YOLOv7 and YOLOv8 are highly suitable for plant disease detection due to several key factors that enhance their performance in real-time, accuracy, and efficiency. Firstly, their speed and efficiency in inference are critical for real-time applications in agriculture, where quick decision-making is essential for managing plant health. YOLOv7 and YOLOv8 are optimized for fast object detection, with inference speeds as low as 3.8ms, enabling swift identification of disease symptoms on plant leaves without significant delays.

Secondly, these models benefit from state-of-the-art architectures that are fine-tuned for object detection tasks. YOLOv7 and YOLOv8 employ advanced techniques such as improved backbone networks and more efficient feature extraction methods, which enhance their ability to detect complex patterns and subtle disease symptoms in plant images. This makes them highly accurate, as demonstrated by their high mAP scores (YOLOv7: 86.3%, YOLOv8: 91.05%), which reflect their ability to correctly classify and localize plant diseases.

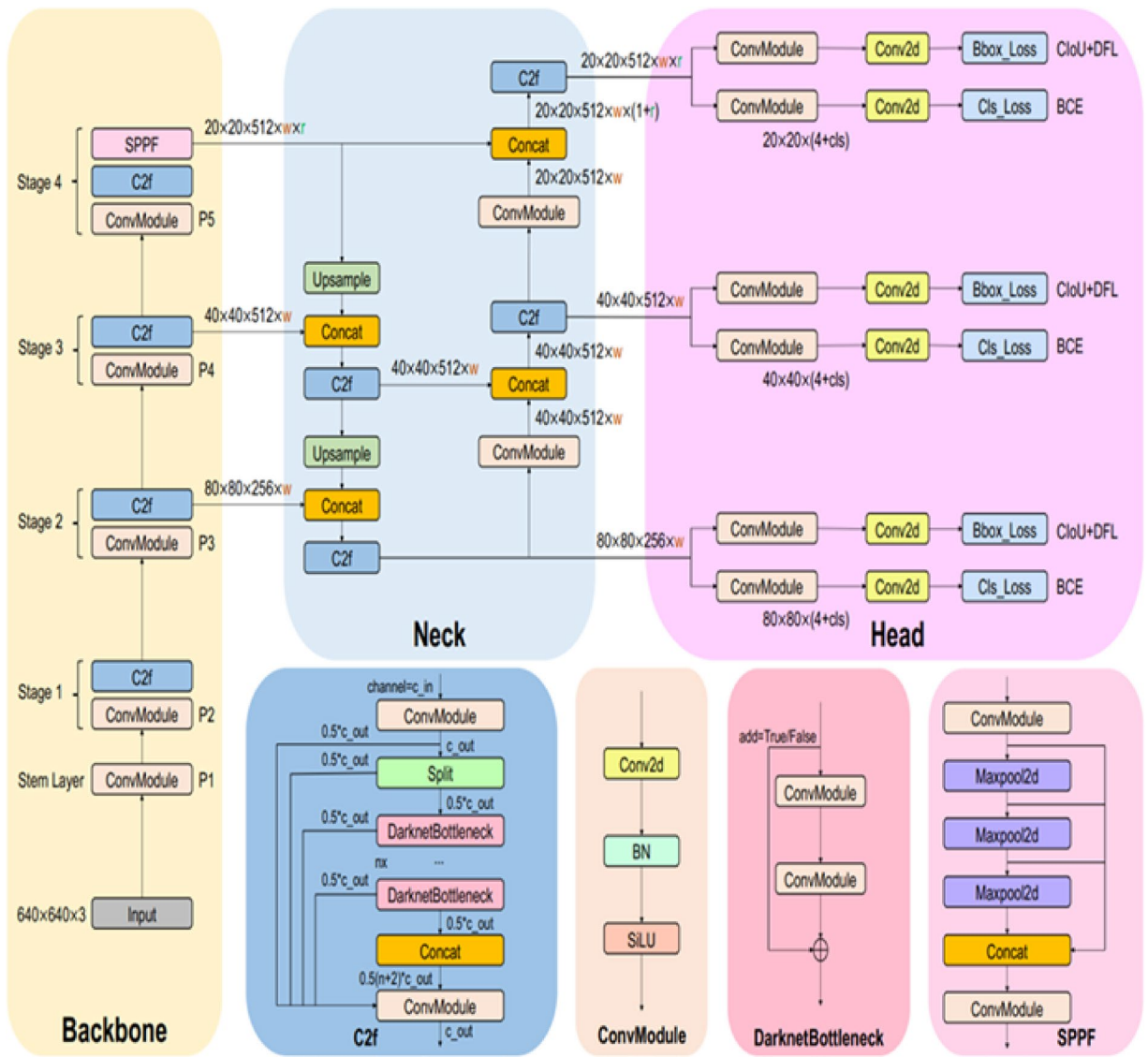

**Fig. 3.** Architecture of YOLOv8.

Another factor contributing to their suitability is their scalability and flexibility. Both models can be trained on various datasets, making them adaptable to different plant species and types of diseases. The flexibility to integrate additional layers or transfer learning techniques allows for fine-tuning, improving performance even with smaller, domain-specific datasets. This adaptability is especially important in agriculture, where different environmental conditions and plant varieties may present unique challenges.

Moreover, YOLOv7 and YOLOv8's robustness to diverse input data further enhances their applicability to plant disease detection. These models excel in handling variations in lighting, background noise, and image resolution, which are common issues when capturing images in agricultural fields. Their ability to detect diseases under varied conditions ensures consistent performance in real-world scenarios.

## Experimental environment and setup

The experimental setup for the plant disease detection system was designed to efficiently train and evaluate deep learning models, particularly for real-time disease detection in Tomato plant leaves. This setup was critical in ensuring both high performance and accuracy, as deep learning models require significant computational resources for training and inference. Below is a detailed explanation of the environment and the various components that contributed to the overall setup.

## Hardware configuration

To handle the intensive computational demands of deep learning, the system was built around high-performance hardware. The core component of the setup was a server equipped with an NVIDIA Tesla V100 or RTX 3090

GPU. These GPUs are specifically designed for machine learning tasks and provide the computational power required to train large models in a reasonable timeframe. The GPU accelerated both the training and inference processes, allowing for faster model convergence and real-time detection of plant diseases. Alongside the GPU, the system had 32GB of RAM to handle the large datasets and the memory-intensive operations during training. For storage, 1 TB SSDs were used to store the training datasets and pre-trained model weights, ensuring fast data read and write speeds during the experimentation process.

## Software environment

The software environment was based on Ubuntu 20.04 LTS, a Linux distribution known for its stability and efficiency when running machine learning applications. Python was chosen as the primary programming language due to its wide range of libraries and frameworks that support machine learning and computer vision tasks. The TensorFlow and PyTorch libraries were employed for building, training, and fine-tuning deep learning models, specifically YOLOv7 and YOLOv8, for plant disease detection. These frameworks offer powerful tools for designing and implementing deep learning models, including pre-built layers for convolutional neural networks (CNNs) and object detection.

Additionally, the environment utilized Keras, a high-level neural networks API, for model building and training. To handle image pre-processing tasks, OpenCV was used for resizing, cropping, and augmenting images, while Matplotlib was used for data visualization, including plotting training loss curves and evaluating model performance. All software dependencies were installed and managed using Docker, creating isolated containers for reproducibility and version control. Docker ensures that the environment is consistent across different machines and experiment runs, which is crucial for maintaining accuracy and reliability.

## Dataset preparation

The dataset used for training and evaluating the plant disease detection models consisted of high-resolution images of plant leaves, each labeled with disease information. A combination of publicly available datasets, such as Tomato trees from PlantVillages, and custom-collected Tomato leaf images were used to ensure diversity in plant species and diseases such as, Powdery Mildew, Angular Leaf Spot, Early blight and Tomato mosaic virus. The dataset contained multiple classes of diseases, each representing a specific plant condition. Prior to training, the images were preprocessed to standardize the input data. This included resizing all images to a consistent resolution (e.g., 224×224 or 256×256 pixels), as input size consistency is crucial for model performance. The pixel values of the images were normalized to a range between 0 and 1 to facilitate faster convergence during training.

To further improve model generalization, data augmentation techniques were applied. This included random rotations, flipping, and scaling of tomato images, which simulated various environmental conditions and tomato plant orientations. Augmentation helps the model become more robust by allowing it to learn from a wider variety of inputs. Additionally, images were divided into training, validation, and test sets, ensuring that the model was evaluated on unseen data to prevent overfitting.

## Model architecture and training process

The key focus of the experiments was to fine-tune YOLOv7 and YOLOv8, which are state-of-the-art models for object detection. These models are specifically designed for real-time applications and can efficiently detect plant diseases by classifying diseases in images and pinpointing their locations. Initially, pre-trained weights from large-scale datasets such as ImageNet were used, leveraging transfer learning. This approach is beneficial as it allows the model to benefit from the features learned by pre-trained models on vast datasets, even if only a smaller, domain-specific dataset is available for fine-tuning.

The training process involved adjusting several key hyperparameters to ensure optimal performance. These included the learning rate, batch size, and the number of epochs. A learning rate schedule was employed to reduce the learning rate gradually as the model converged[14]. The batch size was set to 32 to balance between computational efficiency and model accuracy. The training ran for 50–100 epochs, depending on model convergence, with frequent checkpoints to save intermediate results.

A key part of training was model regularization to avoid overfitting. Techniques such as dropout and early stopping were employed to ensure that the model learned generalizable features rather than memorizing the training data. Additionally, Adam optimizer was used for gradient descent optimization due to its efficiency in handling sparse gradients and adaptive learning rates.

## Performance evaluation

The evaluation of the trained models was carried out using several standard metrics to measure their accuracy and effectiveness in detecting plant diseases. These metrics included:

- Precision: The ratio of true positive predictions to the total number of positive predictions made. It evaluates how many of the predicted diseased plants were correctly identified.
- Recall: The ratio of true positive predictions to the total number of actual diseased instances. It measures how many of the actual diseased plants were correctly detected.
- F1-score: The harmonic mean of Precision and Recall, providing a balanced measure of the model's performance, especially in imbalanced datasets.
- Mean average precision (mAP): A metric used for object detection tasks, mAP summarizes the overall precision at various recall levels across different classes. It is commonly used to assess how well a model performs in localizing and classifying objects.

The model was validated using a separate validation dataset to prevent overfitting to the training data, and performance was tested on a test set to evaluate generalization. The results were compared across different models, and the final model with the highest accuracy, F1-score, and mAP was selected for deployment.

## Results and discussion

The predictions generated by the classification methods are summarized in the confusion matrix. The classification technique's confusion matrix displays the values for true negatives (TN), true positives (TP), false positives (FP), and false negatives (FN).

$$\text{Precision} = \frac{TP}{TP + FP} \tag{1}$$

$$\text{Recall} = \frac{TP}{TP + FN} \tag{2}$$

$$\text{F1Score} = 2 \times \frac{\text{Recall} \times \text{Precision}}{\text{Recall} + \text{Precision}} \tag{3}$$

The prediction of the bounding box's accuracy is evaluated by intersection over union (IoU). The equation below calculates the intersection over union (IOU), which represents the proportion of the area shared by the predicted bounding box (PBB) and the ground truth bounding box (TBB) in the sample images.

$$IoU = \frac{\text{area}(\text{PBB} \cap \text{TBB})}{\text{area}(\text{PBB} \cup \text{TBB})} \tag{4}$$

The average of the precision-recall curve's area under a specific IoU threshold is known as the average precision at IOU ($AP^{IOU}$)[18]. $AP^{IOU}$ is a performance indicator for a certain class or category n. In order to indicate the performance for all detection classes, mean average precision at a threshold IoU ($mAP^{IOU}$) is calculated and denoted as follows:

$$mAP^{IoU} = \frac{1}{N} \sum AP_n^{IoU}; \ n \in \{class1, class2, \ldots, classN\} \tag{5}$$

The effectiveness of the YOLOv7 and YOLOv8 models is assessed using the Detecting Diseases dataset. Results indicate that YOLOv8 achieves a higher mean Average Precision (mAP) and faster inference speed compared to YOLOv7 as illustrated in (Table 1).

The performance of several models was evaluated using the Disease Detection and PlantDoc datasets, focusing on their accuracy and inference speed. MobileNet, tested on the PlantDoc dataset, achieved a mean Average Precision (mAP) of 32.8%, while Faster-RCNN with Inception-ResNet reached an mAP of 38.9%. Specifically, MobileNet was selected for its lightweight architecture and efficiency on mobile and embedded devices, aligning with our goal of enabling real-time plant disease detection in resource-constrained environments. Faster R-CNN, on the other hand, serves as a strong two-stage detector benchmark known for high detection accuracy, albeit with higher computational costs. In contrast, YOLO models are single-stage detectors that prioritize speed and real-time performance, sometimes at the expense of slight accuracy loss. Including MobileNet and Faster R-CNN provides a balanced comparison across different model families (lightweight, two-stage, and single-stage), strengthening the comprehensiveness of our evaluation. In comparison, the proposed YOLOv7 model demonstrated a substantial increase in accuracy, achieving an mAP of 86.3% on the Disease Detection dataset, with a fast inference speed of 4.3 milliseconds. The proposed YOLOv8 model outperformed YOLOv7, achieving a higher mAP of 91.05% and an even faster inference time of 3.8 milliseconds. These results highlight the superior performance of YOLOv7 and YOLOv8 in both accuracy and processing speed (Table 2), making them highly suitable for real-time plant disease detection in agricultural applications. The study affirms the positions of similar studies[25-32] regarding the efficiency of models in disease prediction and management.

The YOLOv7 and YOLOv8 models were evaluated using randomly selected testing images, and the outcomes are presented in (Fig. 4). The experimental findings demonstrate the successful identification of plant diseases by the models.

The performance of the YOLOv7 and YOLOv8 models was assessed using key metrics, including mAP, F1 score, precision, and recall. After 50 epochs of training, YOLOv8 achieved a mAP of 86.3%, an F1 score of 83.1%, precision of 81.9%, and recall of 84.3%. In comparison, YOLOv7 outperformed YOLOv8, reaching a mAP of 91.05%, an F1 score of 89.40% (Fig. 5), precision of 91.22%, and recall of 87.66% (Fig. 6). The corresponding performance graphs are presented in (Figs. 7, 8, 9, 10, 11, 12, 13 and 14).

| Model | Dataset | mAP (at 50% iou) | Inference speed |
|---|---|---|---|
| MobileNet[1] | PlantDoc | 32.8 | – |
| Faster-rcnn-inception-Resnet[1] | PlantDoc | 38.9 | – |
| **Proposed YOLOv7** | **Disease Detection** | **86.3** | **4.3ms** |
| **Proposed YOLOv8** | **Disease Detection** | **91.05** | **3.8ms** |

**Table 1.** Comparison of leaf detection mAP.

| S.no | Classtives | True positive (TP) | False positive (FP) | False negative (FN) | True negative (TN) | Precision | Recall | F1-score | Accuracy |
|------|-----------|--------------------|--------------------|--------------------|--------------------|-----------|--------|----------|----------|
| 0 | Class 1 | 50 | 1 | 1 | 198 | 0.98 | 0.98 | 0.98 | 0.94 |
| 1 | Class 2 | 45 | 6 | 6 | 195 | 0.88 | 0.88 | 0.88 | 0.90 |
| 2 | Class 3 | 52 | 0 | 1 | 49 | 1.00 | 0.98 | 0.99 | 0.95 |
| 3 | Class 4 | 46 | 5 | 3 | 46 | 0.90 | 0.94 | 0.92 | 0.90 |
| 4 | Class 5 | 48 | 2 | 2 | 50 | 0.96 | 0.96 | 0.96 | 0.94 |

**Table 2.** Comprehensive performance summary on various training and validation stages.

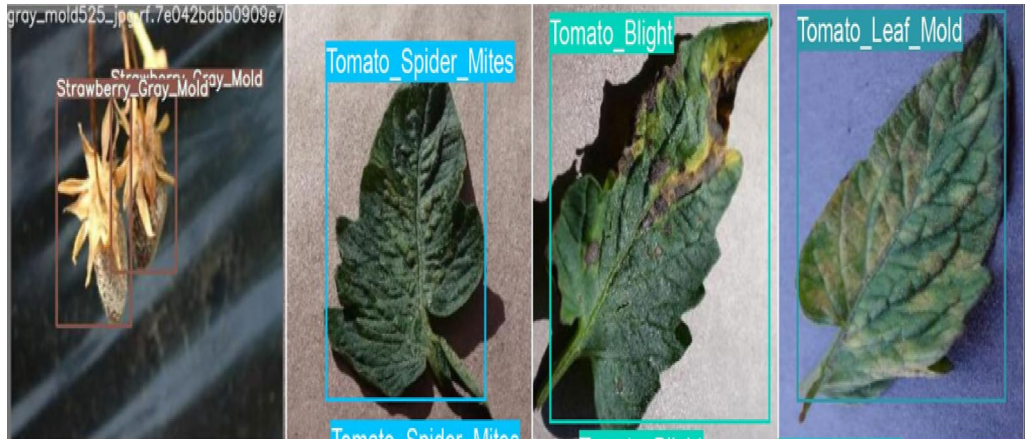

**Fig. 4.** Detection of a specific class on sample test images (tomato leaf).

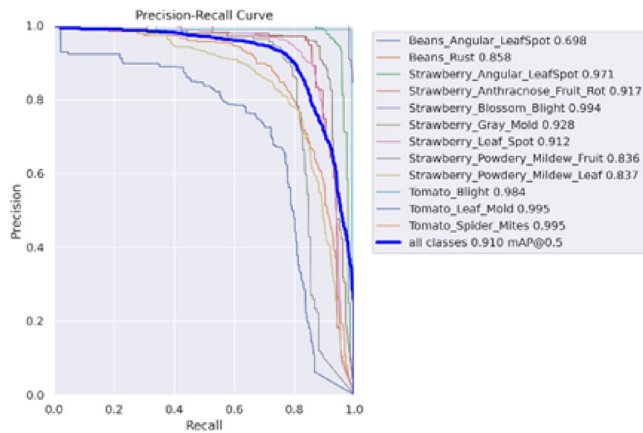

**Fig. 5.** Precision-recall for YOLOv8.

The findings from applying transfer learning and deep learning techniques to plant disease detection have significant real-world implications, such as Enhancing Agricultural Productivity and Efficiency, Support for Small-Scale and Resource-Constrained Farmers, Impact on Global Food Security, Environmental Sustainability and Reduced Chemical Usage, Potential for Integration with Other Technologies, Advancing Research and Innovation in Agriculture, Capacity for Climate Change Adaptation, etc. Particularly in enhancing agricultural efficiency and sustainability. By enabling early and accurate identification of plant diseases, these technologies can help farmers reduce crop loss, minimize pesticide use, and optimize resource allocation, leading to cost savings and improved yields. This not only supports food security by ensuring healthier crops but also promotes environmentally responsible farming practices. Moreover, the accessibility of these solutions through mobile devices offers small-scale and resource-constrained farmers in remote areas the opportunity to adopt advanced, low-cost diagnostic tools. Ultimately, these advancements have the potential to revolutionize agricultural practices, contributing to more sustainable, resilient, and productive global food systems.

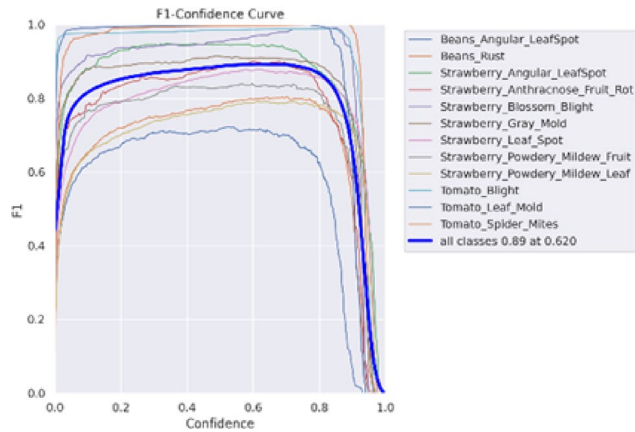

**Fig. 6.** F1score for YOLOv8.

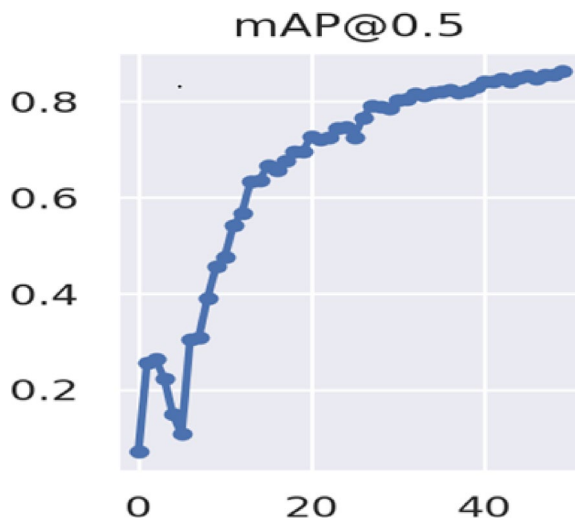

**Fig. 7.** mAP for YOLOv7.

In terms of speed, accuracy, and computing efficiency, YOLOv7 and YOLOv8 beat existing object identification models for plant disease diagnosis, including Faster R-CNN and Efficient Det. On datasets such as PlantVillage and Kaggle's plant pathogen dataset, YOLOv7 and YOLOv8 offer quick real-time processing, making them appropriate for practical applications such as drone-based or mobile plant monitoring for diseases. While models such as Faster R-CNN and the other related models such as the one by Uskaner[33] are very accurate, they are typically slower due to more complicated structures and multi-stage processing. YOLO models provide a balance between high detection accuracy and low computing requirements, allowing implementation on devices with restricted resources without sacrificing performance. This efficiency and scalability make YOLOv7 and YOLOv8 ideal for large-scale agricultural surveillance.

## Practical implementations and integration of the proposed model

The proposed solution can be integrated into mobile applications for farmers, incorporated into drone-based disease monitoring systems to broaden its impact on precision agriculture practices. It can be successfully included into farmer mobile applications by providing real-time, easily available diagnostic tools that employ environmental data and picture recognition to identify early indicators of crop illnesses. It can also be included into drone-based monitoring systems, in which high-resolution camera and sensor drones fly over fields, and AI onboard or in the cloud evaluates the data to rapidly spot disease trends over wide regions. By facilitating prompt interventions, reducing needless pesticide usage, increasing crop yields, and promoting more sustainable farming methods, these applications taken together significantly enhance precision agriculture techniques. By providing farmers with actionable insights, this combination of predictive technology and intelligent deployment might hasten the shift to more effective, data-driven agricultural management.

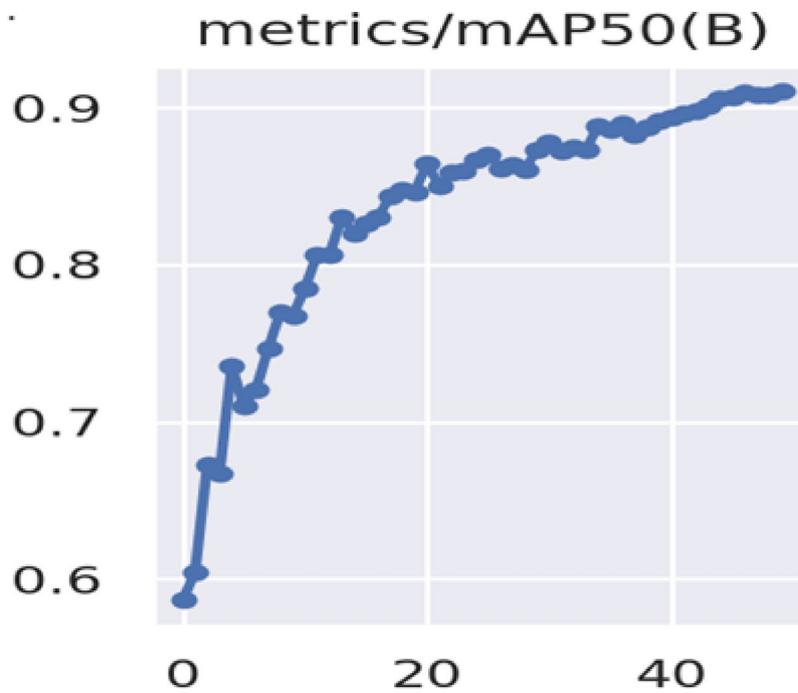

**Fig. 8**. mAP for YOLOv8.

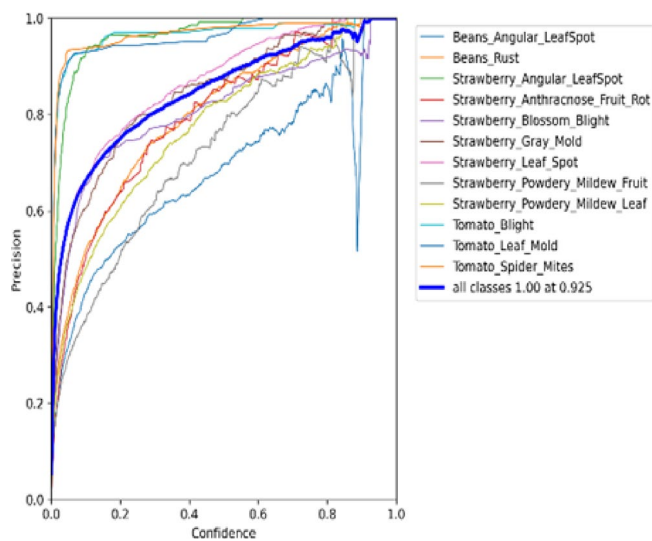

**Fig. 9**. Precision for YOLOv7.

## Conclusion

The process of identifying and categorizing plant diseases through digital imagery poses significant challenges. Consequently, timely and accurate disease identification is highly important for farmers and plant pathologists to respond effectively. The proposed method utilizes a Detecting Diseases dataset containing images representing 12 distinct plant disease types. YOLOv7 and YOLOv8 object detection models are trained on this dataset. Notably, the proposed YOLOv8 model achieves an impressive mean Average Precision (mAP) score of 91.05%, highlighting its effectiveness in accurately identifying and classifying plant diseases. This approach has the potential to predict diseases early, mitigate crop losses, and enhance economic gains in the agricultural sector.

The study helps advance the field by demonstrating a reliable and extensible method for detecting plant diseases, which takes advantage of transfer learning's characteristics to improve accuracy and usefulness across a wide range of agricultural settings. The work shows how pre-trained models can be used to detect plant diseases without requiring a large amount of training data or computer resources. This strategy makes use of existing information from models trained on huge datasets to improve plant disease detection efficiency and accuracy.

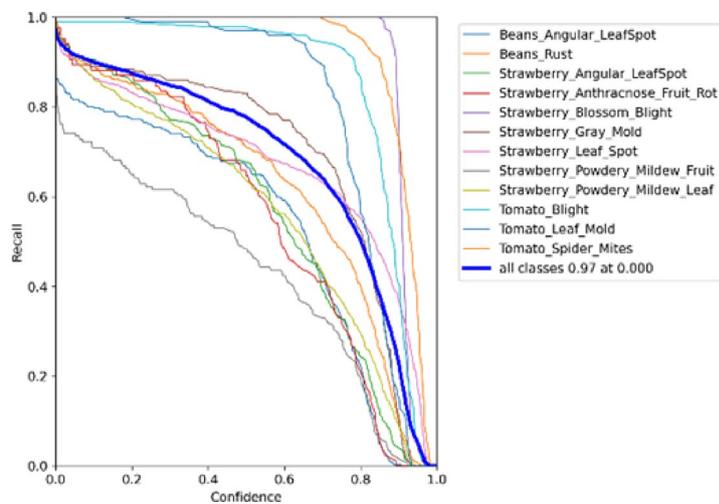

**Fig. 10**. Recall for YOLOv7.

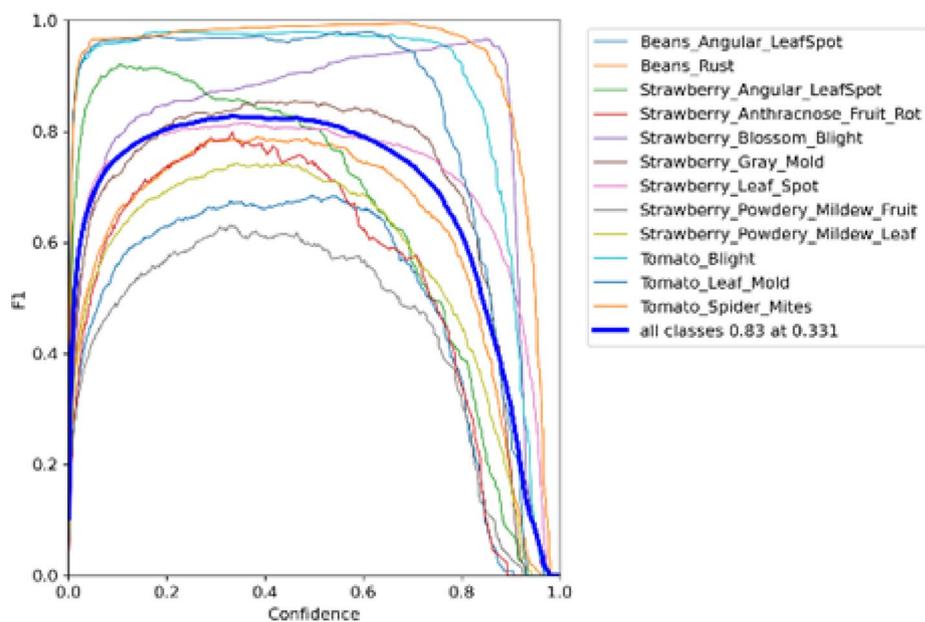

**Fig. 11**. F1score for YOLOv7.

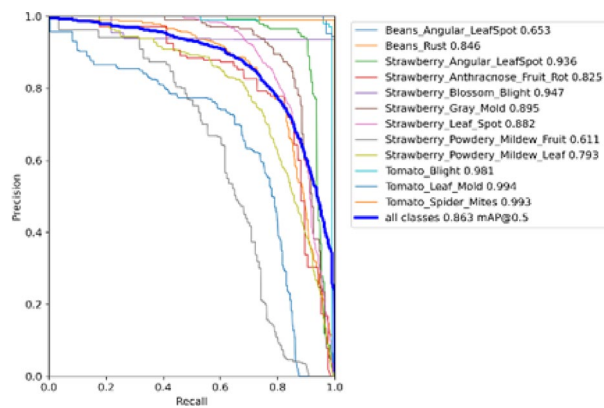

**Fig. 12**. Precision-recall for YOLOv7.

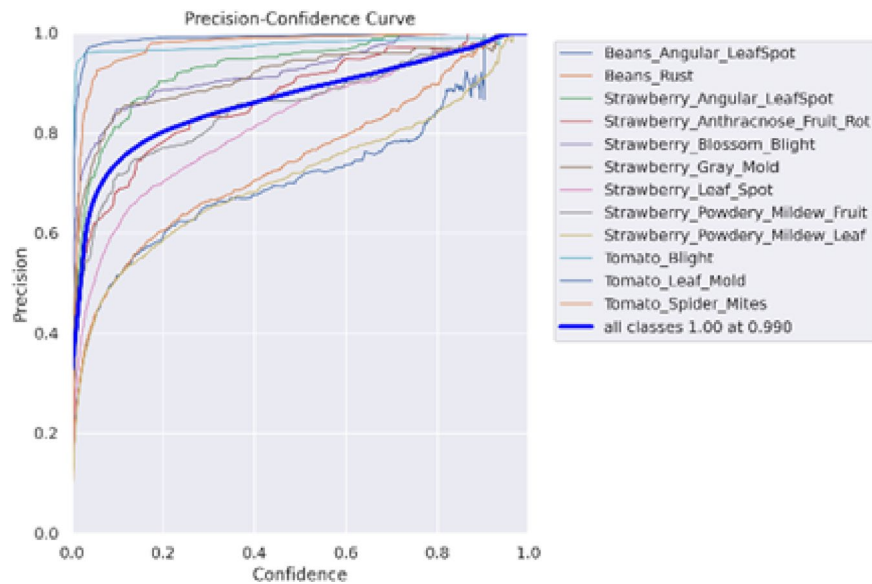

**Fig. 13**. Precision for YOLOv8.

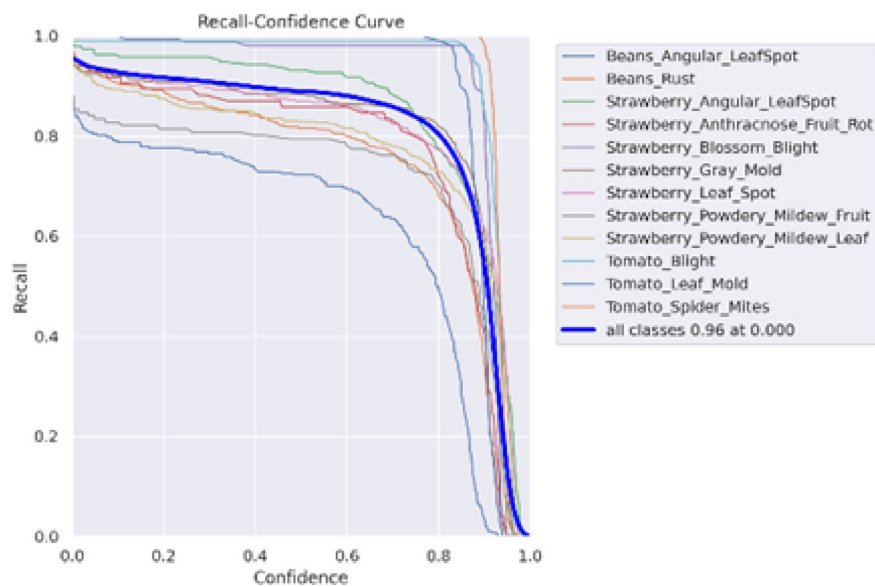

**Fig. 14**. Recall for YOLOv8.

In future work, the plant leaf dataset can be expanded by including more images of various plant leaves. This expansion aims to improve the models' predictive accuracy, especially in challenging situations. Subsequent studies in this field can emphasize the refinement of algorithms to enhance the efficiency and effectiveness of detecting diseased leaves.

## Data availability
The data can be accessed via: https://drive.google.com/file/d/1kA_JWhHQhyzzuzlpzppK2nNTtbiR2N77/view. https://github.com/Sachinthana-Lokuyaddage/Plant_Disease_Detection_Using_Transfer_Learning_with_ResNet50.

## References

1. Li, C. et al. Yolov6: A single-stage object detection framework for industrial applications. arXiv.

2. Piotr Doll´ar, M., Singh & Girshick, R. Fast and accurate model scaling. In *Proceedings of the IEEE/CVF Conference on Computer Vision and PatternRecognition* 924–932, (2021).

3. Wajid, M. A., Zafar, A., Wajid, M. S. & Terashima-Marín, H. Neutrosophic-CNN-based image and text fusion for multimodal classification. *J. Intell. Fuzzy Syst.*, 1–17 .

4. Singh, D. et al. PlantDoc: a dataset for visual plant disease detection. In *Proceedings of the 7th ACM IKDD CoDS and 25th COMAD* (pp. 249–253) (2020).

5. Kasso, M. & Bekele, A. Post-harvest loss and quality deterioration of horticultural crops in dire Dawa region, Ethiopia. *J. Saudi Soc. Agricultural Sci.* **17** (1), 88–96 (2018).

6. Almeida, R. P. Emerging plant disease epidemics: biological research is key but not enough. *PLoS Biol.* **16** (8), e2007020 (2018).

7. Fang, Y. & Ramasamy, R. P. Current and prospective methods for plant disease detection. *Biosensors* **5** (3), 537–561 (2015).

8. Marwan Adnan Jasim and Jamal Mustafa Al-Tuwaijari. Plant leaf diseases detection and classification using image processing and deep learning techniques. In *2020 International Conference on Computer Science and Software Engineering (CSASE)*. 259–265. (IEEE, 2020).

9. Mohanty, S. P., Hughes, D. P. & Salathé, M. Using deep learning for image-based plant disease detection. *Front. Plant Sci.* **7**, 1419 (2016).

10. Wang, C. Y., Bochkovskiy, A. & Liao, H. Y. YOLOv7: trainable bag-of-freebies sets new state-of-the-art for real-time object detectors. *ArXiv* (2022)..

11. Haque, M. E., Rahman, A. & Junaeid, I. Samiul Ul Hoque, and Manoranjan Paul. Rice leaf disease classification and detection using yolov5. *arXiv*, arXiv–2209, (2022).

12. Pavan Kumar Anasosalu Vasu & Gabriel, J. *Jeff Zhu, OncelTuzel, and Anurag Ranjan. An Improved One Millisecond Mobile Backbone* arXiv–2206 (arXiv e-prints, 2022).

13. Gao, P., Lu, J. & Li, H. Roozbeh Mottaghi, and Aniruddha Kembhavi. Container: context aggregation network. *ArXiv* (2021).

14. Wajid, M. S., Terashima-Marin, H., Rad, P., Wajid, M. A. & P. N., & Violence detection approach based on cloud data and neutrosophic cognitive maps. *J. Cloud Comput.* **11** (1), 1–18 (2022).

15. Wajid, M. S. & Wajid, M. A. The importance of indeterminate and unknown factors in nourishing crime: A case study of South Africa using neutrosophy. *Neutrosophic Sets Syst.* **41**, 15 (2021).

16. Mohapatra, T. ICAR News July-September 2018. Published in the monthly newsletter. https://www.icar.org.in/sites/default/files/ICARNewsJulySeptember2018.pdf (2018).

17. Zafar, A. & Wajid, M. A. *Neutrosophic Cognitive Maps for Situation Analysis* (Infinite Study, 2019).

18. Wajid, M. S., Maurya, S. & Vaishya, R. Sentence similarity-based text summarization using clusters. *Int. J. Sci. Eng. Res.* 4. (2013).

19. Lee, S. H., Goëau, H., Bonnet, P. & Joly, A. New perspectives on plant disease characterization based on deep learning. *Comput. Electron. Agric.* **170**, 105220 (2020).

20. Artificial Intelligence. Detecting Diseases Dataset. RoboflowUniverse, Rob flow. https://universe.roboflow.com/artificial-intelligence-82oex/detecting-diseases (2022).

21. Hildreth, J. et al. Standard operating procedure for the collection and Preparation of voucher plant specimens for use in the nutraceutical industry. *Anal. Bioanal Chem.* **389**, 13–17. https://doi.org/10.1007/s00216-007-1405-x (2007).

22. Sujatha, R., Chatterjee, J. M., Jhanjhi, N. Z. & Brohi, S. N. Performance of deep learning vs machine learning in plant leaf disease detection. *Microprocess. Microsyst.* **80**, 103615 (2021).

23. Narayana, C. L. Kondapalli Venkata Ramana. An efficient real-time weed detection technique using YOLOv7. *Int. J. Adv. Comput. Sci. Appl.* **14**, 2 (2023).

24. Midhun, P. & Mathew and Therese Yamuna Mahesh. Leaf-based disease detection in bell pepper plant using Yolo v5. *Signal Image Video Process.* 1–7, (2022).

25. Sharma, H. et al. Multi-modal data fusion using transfer learning in big data analytics for healthcare. In *International Conference on Artificial Intelligence for Innovations in Healthcare Industries, ICAIIHI* https://doi.org/10.1109/ICAIIHI57871.2023.10489309 (2023).

26. Deepak, A. et al. Image processing based robotic car for agricultural ploughing using machine learning approach. *Int. J. Intell. Syst. Appl. Eng.* **12**, 2s718 (2024).

27. Sinha, A. et al. Embodied Understanding of large Language models using calibration enhancement. *Int. J. Intell. Syst. Appl. Eng.* **12** (13s), 59–66 (2024).

28. Mishra, J. S. et al. Evaluating the effectiveness of heart disease prediction. *Int. J. Intell. Syst. Appl. Eng.* **12** (5s), 163–173 (10).

29. Bhadula, S. et al. Time series analysis for power grid anomaly detection using LSTM networks. *Proc. Int. Conf. Communication Comput. Sci. Eng. IC3SE* **1358-1363** (5). https://doi.org/10.1109/IC3SE62002.2024.10593319 (2024).

30. Shankar, G. S., Onyema, E. M., Kavin, B. P., Gude, V. & Prasad, B. S. Breast cancer diagnosis using virtualization and extreme learning algorithm based on deep feed forward networks. *Biomedical Eng. Comput. Biol.* **15** https://doi.org/10.1177/11795972241278907 (2024).

31. Ugboaja, S. G. et al. Advanced diabetes prediction using supervised machine learning technique: randomforest. *Trop. J. Appl. Nat. Sci.* **2** (3), 1–14 (2024).

32. Selvaraj, S. et al. Uchechi, AQ. super learner model for classifying leukemia through gene expression monitoring. *Discover Oncol.* **15**, 499. https://doi.org/10.1007/s12672-024-01337-x (2024).

33. Uskaner, H. P. Efficient plant disease identification using few-shot learning: a transfer learning approach. *Multimed. Tools Appl.* **83**, 58293–58308. https://doi.org/10.1007/s11042-023-17824-2 (2024).

## Acknowledgements
We thank everyone who inspired our work.

## Author contributions
BSM: conceptualization and drafting, HSN: data collection and analysis. MNW: Methodology and review. NOF: Drafting and interpretation. CCZ: Writing and method. HDA: writing and Data Collection; EMO: Drafting, Supervision and Editing.

## Declarations

## Competing interests
The authors declare no competing interests.

## Consent to participate
All authors consent to participate.

## Additional information

**Correspondence** and requests for materials should be addressed to N.O.F.

**Reprints and permissions information** is available at www.nature.com/reprints.

**Publisher's note** Springer Nature remains neutral with regard to jurisdictional claims in published maps and institutional affiliations.